\documentclass[10pt,twocolumn,letterpaper]{article}
\usepackage[accsupp]{axessibility}
\usepackage[final]{cvpr}      

\usepackage[dvipsnames]{xcolor}

\definecolor{cvprblue}{rgb}{0.21,0.49,0.74}
\usepackage[pagebackref,breaklinks,colorlinks,citecolor=cvprblue]{hyperref}

\def\paperID{5} 
\def\confName{CVPRW}
\def\confYear{2024}

\usepackage{subfloat}
\usepackage{booktabs}
\usepackage{diagbox}
\usepackage{multirow}
\usepackage{multicol}
\usepackage{makecell}
\usepackage{amsmath}
\usepackage{algorithm}
\usepackage{pifont}
\newcommand{\cmark}{\ding{51}}%
\newcommand{\xmark}{\ding{55}}%
\usepackage{algpseudocode}
\usepackage{cuted}

\title{Source-free Domain Adaptation for Video Object Detection \\
Under Adverse Image Conditions}

\author{Xingguang Zhang\\
Purdue University\\
West Lafayette, In 47907, USA\\
{\tt\small zhan3275@purdue.edu}
\and
Chih-Hsien Chou\\
Futurewei Technologies, Inc.\\
Santa Clara CA 95050, USA\\
{\tt\small cchou@futurewei.com}
}

\begin{document}
\maketitle
\begin{abstract}
When deploying pre-trained video object detectors in real-world scenarios, the domain gap between training and testing data caused by adverse image conditions often leads to performance degradation. Addressing this issue becomes particularly challenging when only the pre-trained model and degraded videos are available. Although various source-free domain adaptation (SFDA) methods have been proposed for single-frame object detectors, SFDA for video object detection (VOD) remains unexplored. Moreover, most unsupervised domain adaptation works for object detection rely on two-stage detectors, while SFDA for one-stage detectors, which are more vulnerable to fine-tuning, is not well addressed in the literature. In this paper, we propose Spatial-Temporal Alternate Refinement with Mean Teacher (STAR-MT), a simple yet effective SFDA method for VOD. Specifically, we aim to improve the performance of the one-stage VOD method, \emph{YOLOV}, under adverse image conditions, including \emph{noise}, \emph{air turbulence}, and \emph{haze}. Extensive experiments on the ImageNetVOD dataset and its degraded versions demonstrate that our method consistently improves video object detection performance in challenging imaging conditions, showcasing its potential for real-world applications.

\end{abstract}    
\section{Introduction}
\label{sec:intro}
\begin{figure}
    \centering
    \includegraphics[width = 0.95\linewidth]{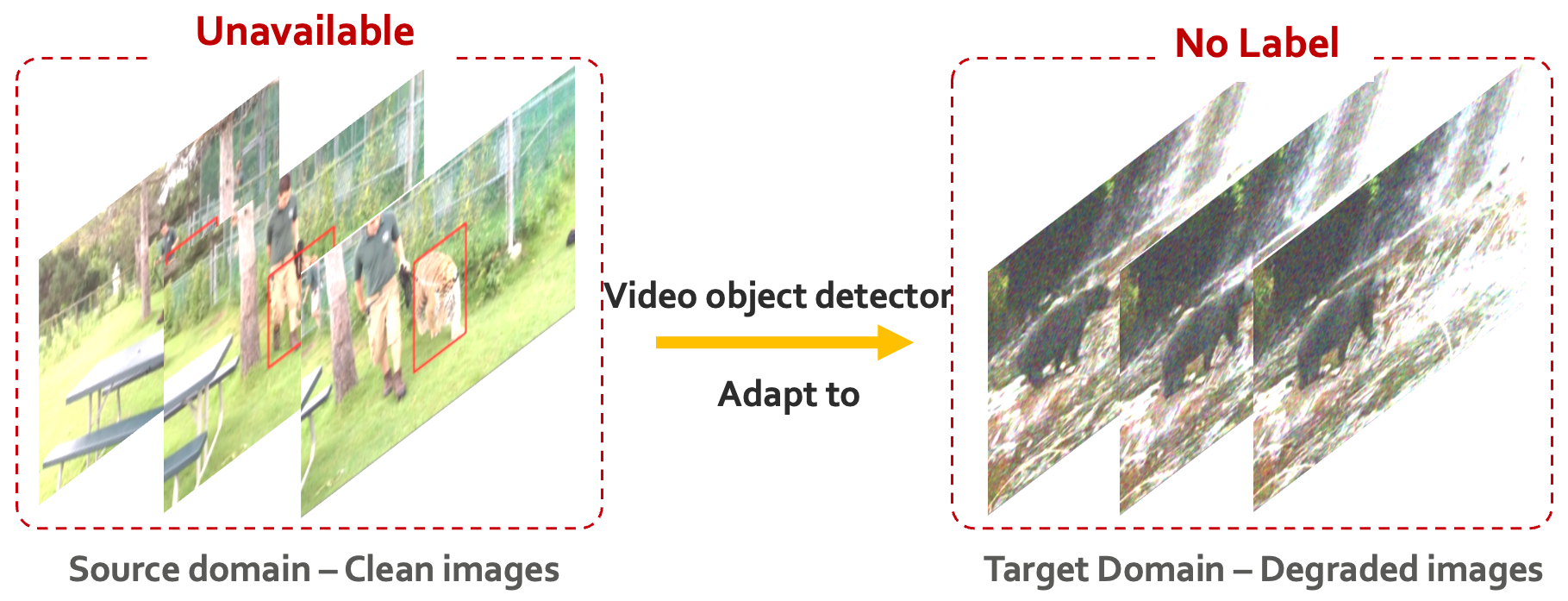}
    \caption{The scope of this work: we aim to adapt the video object detection model trained on clean image sequences to degraded image sequences under the condition that the data from the source domain and ground truth labels of the target domain are unavailable during the adaptation.}
    \label{fig:overview}
\end{figure}

Object detection in images and videos represents a pivotal task in computer vision, primarily owing to its extensive range of applications across diverse scenarios, such as intelligent surveillance systems and automated driving vehicles. Video object detection (VOD) aims to predict the bounding box and category information of all targeting objects in all video frames. Compared to single-frame object detection tasks, VOD enjoys the advantage of accessing additional information from the temporal dimension \cite{zhu2017flow, shi2023yolov}, which often contains consistent semantics and multiple views of the same target to help enrich the feature space and facilitate superior performance.

When deploying object detectors in real-world settings, adverse imaging conditions caused by rain, haze, low light, or air turbulence often lead to a notable domain gap. These adverse conditions frequently result in a significant reduction in performance, underscoring the necessity for domain adaptation to bridge the gap. Typical domain adaptation (DA) methods require access to data and labels in both source and target domains, with labels in the target domain being relatively scarce \cite{adp_vp}. Due to the prohibitive cost of labeling target domain data, unsupervised domain adaptation (UDA) is sometimes required to facilitate effective fine-tuning on the target domain without any labels \cite{oza2023unsupervised}. In most DA and UDA cases for image detection, source domain data is provided to serve as anchors for adaptation. However, in some UDA scenarios, source data may not be available due to storage or privacy constraints, necessitating the optimization of detection networks under such limitations. To address this challenge, source-free domain adaptation (SFDA) is gaining increasing attention in the field of object detection, as it enables adaptation to target domains without relying on source domain data. 

\begin{figure*}
    \centering
    \includegraphics[width = 0.95\linewidth]{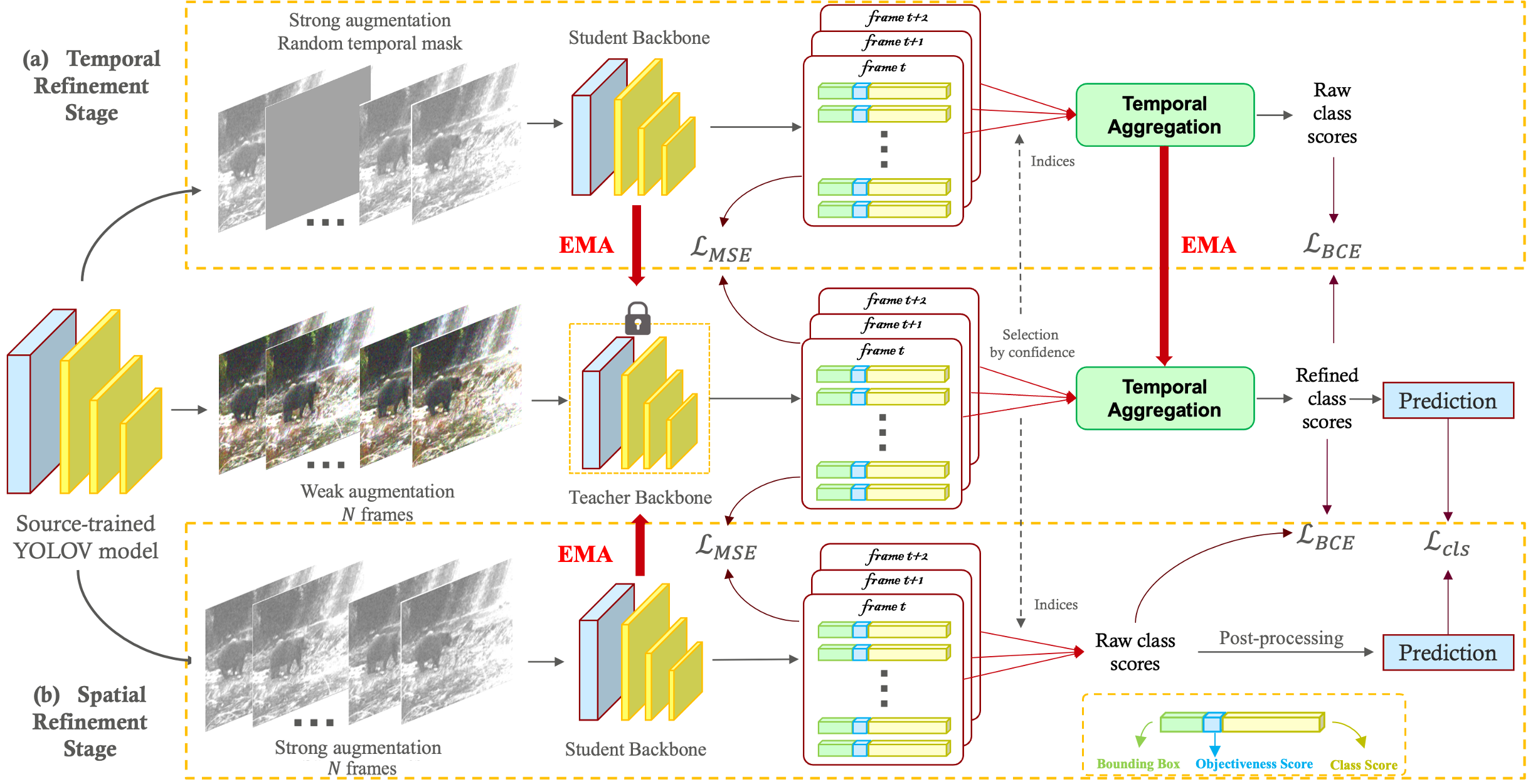}
    \caption{Overview of the proposed STAR-MT for source-free adaptive video object detection. The domain adaptive fine-tuning alternately operates in two stages: (a) Temporal Refinement Stage (TRS) and (b) Spatial Refinement Stage (SRS).}
    \label{fig:SFVOD}
\end{figure*}

Although multiple approaches, such as those proposed in \cite{vibashan2023instance, li2022source, run_and_chase, chen2023exploiting, liu2023periodically, li2021free, cao2023contrastive}, have been developed recently to address SFDA for object detection, they predominantly focus on two-stage object detectors \cite{faster_rcnn}. To date, no prior work has specifically targeted SFDA for one-stage object detectors \cite{YOLOv1}, as their weights are more sensitive to fine-tuning, and intermediate features are highly abstract. Moreover, no SFDA method has been proposed for the video object detection task, primarily due to the scarcity of cross-domain video object detection datasets. However, given the unsupervised nature of SFDA, it is possible to explore source-free domain adaptation methods for video object detection using synthetic data. The scope of this work is illustrated in Fig. \ref{fig:overview}. One can expect those methods to be reliably applied to real-world scenarios where source domain data may not be readily available or accessible.

In this paper, we conduct analysis and experiments about basic SFDA techniques for the video object detection task and propose a novel SFDA method for the one-stage video object detector YOLOV \cite{shi2023yolov}. Specifically, we aim to adapt YOLOV \cite{shi2023yolov} to challenging adverse image conditions by alternately fine-tuning the video object detection model in the teacher-student learning framework. We summarize the contributions of this paper as follows:
\begin{itemize}
    \item We conduct the pioneering study exploring source-free unsupervised domain adaptation for video object detection (VOD). Specifically, we aim to adapt the YOLOV detector to adverse image conditions.
    \item We introduce a novel SFDA algorithm for VOD, termed Spatial-Temporal Alternate Refinement with Mean Teacher (STAR-MT). It alternately trains in two stages: the Temporal Refinement Stage works in a traditional mean-teacher learning scheme, while the Spatial Refinement Stage leverages temporally enhanced features to guide the single-frame backbone module. 
    \item We demonstrate the effectiveness of our method through experiments on different synthetic adverse image conditions, including noise, air turbulence, and haze. Given its unsupervised nature, STAR-MT is anticipated to yield reliable performance boost for video object detectors in real-world and unseen scenarios.

\end{itemize}

\section{Related works}
\label{sec:related}

\subsection{Video object detection.} 
Object detection, one of the most fundamental problems in computer vision, aims to predict the location and class of objects of interest within an input image. Neural networks for object detection can be roughly categorized into one-stage and two-stage detectors \cite{zou2023object}. Two-stage detectors, represented by Faster RCNN \cite{faster_rcnn} and FPN \cite{lin2017feature}, predict the feature proposals and detection results in a two-step, coarse-to-fine manner. These detectors generally exhibit good performance and are relatively easy to train. In contrast, one-stage detectors, such as YOLO \cite{YOLOv1}, SSD \cite{liu2016ssd}, and DETR \cite{carion2020end}, predict all detections in a single inference stage, making them faster and easier to deploy in real-world applications. However, training one-stage detectors often requires more tricks, and they may struggle when detecting dense and small targets.

Video object detection (VOD) presents a unique set of challenges distinct from still image object detection, primarily due to the dynamic nature of video content. In VOD, temporal consistency across frames in the same sequence can be exploited to enhance the robustness of features and reduce the potential ambiguity of the object information in a single image. 
Given an image sequence $\mathbf{I}\in \mathbb{R}^{H\times W \times T}$ where $H$, $W$ and $T$ are image height, image width, and temporal length of the sequence. Most recent VOD algorithms predict the object location $\{y^{loc}\}$ and category $\{y^{cls}\}$ by extracting spatial features via a single-frame backbone and temporal aggregation \cite{wu2019sequence}.
Those solutions are customized for two-stage image object detectors \cite{faster_rcnn, gong2021temporal} or transformers \cite{wang2022ptseformer, VSformer, zhou2022transvod}. These detectors, while effective, often incur high computational costs due to their model size or complex processing pipeline \cite{li2021free}.

In contrast, YOLOV \cite{li2021free} integrates the one-stage object detector YOLOX \cite{ge2021yolox} as its spatial backbone. This configuration benefits from a cost-effective temporal aggregation module, which significantly enhances YOLOX's performance, endowing YOLOV with both superior performance and efficiency. YOLOV's methodology involves selecting key regions from the dense prediction map produced by the detection head, minimizing the processing of numerous low-quality candidates. Furthermore, it assesses the affinity between extracted features from both target and reference frames, facilitating a lightweight feature aggregation process. This strategy presents an efficient alternative to more cumbersome methods, particularly advantageous in scenarios demanding real-time responsiveness.

\subsection{Source-free domain adaptation} 
Domain adaptation for object detection involves data from two domains with different data distributions: the source domain, in which the detector is initially trained, and the target domain, where the detector will be ultimately deployed. Typically, labeling in the target domain is relatively scarce \cite{adp_vp}. In addition to supervised domain adaptation, methods for semi-supervised \cite{inoue2018cross, saito2019semi} and unsupervised domain adaptation \cite{chen2018domain, saito2019strong, cai2019exploring, vs2021mega} for object detection have been studied based on the availability of labels. These methods have achieved significant success in domain adaptation, regardless of whether labeling is available in the target domain data. However, they all require access to source domain data, which may not always be available due to privacy or storage constraints. To address this issue, methodologies for source-free domain adaptation (SFDA) have been recently proposed \cite{liang2020we, tarvainen2017mean, kundu2020universal, yang2021generalized, liu2021source, yang2022attracting, xu2022source}. 

SFDA aims to adapt the detector to the target domain using only the pre-trained model and target domain data, without requiring access to the source domain data, making it a promising approach for real-world applications where source data may be unavailable or inaccessible. Initial SFDA strategies have harnessed self-supervised techniques and pseudo-labeling \cite{liang2020we}. The mean-teacher method \cite{tarvainen2017mean} employs a student-teacher paradigm where the teacher model's parameters are an exponential moving average of the student model's parameters. This approach has shown effectiveness in stabilizing training and improving robustness as the teacher model accumulates and refines knowledge over time, aiding in generating more reliable pseudo-labels. Such methods underscore the essence of SFDA: leveraging target domain intrinsic properties while circumventing the need for source data, thereby aligning domain-specific feature distributions.

Mean-teacher has been a fundamental technique in SFDA for object detection. Most existing works \cite{vibashan2023instance, li2022source, run_and_chase, chen2023exploiting, liu2023periodically, chu2023adversarial, cao2023contrastive} employ the mean-teacher as part of their frameworks. Besides mean-teacher, other methodologies include self-entropy descent and pseudo-label refinement \cite{li2021free}, style enhancement and graph alignment constraint \cite{li2022source}, adversarial alignment \cite{chu2023adversarial}, instance relation graph \cite{vibashan2023instance} and contrastive representation learning \cite{vibashan2023instance, cao2023contrastive}. However, most existing algorithms are designed for two-stage detectors, particularly the Faster RCNN \cite{faster_rcnn}, and cannot be directly applied to the domain adaptation for one-stage detectors such as the YOLO series. This is partially because the region proposals in two-stage detectors could provide high-quality semantic information for additional feature alignment, providing meaningful additional training signals for SFDA. Another reason is that one-stage detectors usually need complicated training tricks; their feature space is more intractable and vulnerable to fine-tuning. Recently, YOLOV \cite{shi2023yolov} provided an efficient feature selection and fusion mechanism for the one-stage detector YOLOX \cite{ge2021yolox} among multiple frames, SFDA for YOLOV would provide valuable experience in both domain adaptation for one-stage detectors and VOD tasks.

\section{Method}

In this section, we detail our STAR-MT method and the domain adaptation benchmark. The overall scheme of the proposed solution is illustrated in Fig.\ref{fig:SFVOD}. 

\subsection{Mean-teacher for domain adaptive VOD}
 In developing our method, we leverage the advanced unsupervised domain adaptation strategies found in the mean-teacher self-training approach \cite{tarvainen2017mean}. We introduce the implementation of this method in this paradigm.
 
 As a class of student-teacher training approach, the mean-teacher method keeps two identical networks: the student network and the teacher network. They are initialized by the weights trained on the source domain. During training, the weights of the teacher model are fixed, while the student model is trained with the supervision signal from the prediction output and features generated from the teacher model. On the other hand, the teacher model takes the exponential moving average (EMA) of consecutive student models for its parameter update:
\begin{equation}
    \theta_{\mathcal{T}}^{t} \leftarrow \alpha \theta_{\mathcal{T}}^{t-1} + (1-\alpha) \theta_{\mathcal{S}}^{t-1},
\end{equation}
where the $\theta_{\mathcal{T}}$ and $\theta_{\mathcal{S}}$ denote the weights of teacher and student models, $t$ denotes the training iteration, and $\alpha \in (0,1)$ is the momentum coefficient which is usually set close to 1 for a smooth temporal ensemble \cite{cao2023contrastive}.
 
\subsection{Spatial-Temporal Alternate Refinement}
YOLOV utilized the pre-trained backbone of YOLOX as its frame-wise feature extractor, followed by feature selection and affinity measurement that identifies features from the same object among frames to guide temporal aggregation. However, training the spatial backbone and temporal aggregation module simultaneously on the video object detection dataset is suboptimal because they require different training schemes. Hence, we propose to adapt the YOLOV in a two-stage alternate optimization manner, consisting of the temporal refinement stage (TRS) and spatial refinement stage (SRS).

\subsubsection{Temporal Refinement Stage (TRS).}
In the TRS, the entire teacher model, including the frame-wise backbone and temporal aggregation module, is updated via EMA. In the beginning, both teacher and student models are initialized the same.
Like a typical mean-teacher-based algorithm, the same image sequences with different augmentations are fed into those models. The teacher model processes the weakly augmented images, and the heavily augmented images are fed into the student model. Moreover, we randomly mask out $r\%$ frames and enforce the student model to produce the same output with fewer frames than the teacher model. This masking mechanism can supposedly enhance the generalization capability of temporal aggregation. The student model is trained by aligning frame-wise features and soft pseudo labels with the features and predictions of the teacher model. The loss in this stage is defined as:
\begin{equation}
    \mathcal{L} = \mathcal{L}_{MSE}(f_{\mathcal{T}}, f_{\mathcal{S}}) + \mathcal{L}_{BCE}(y_{\mathcal{T}}^{cls}, y_{\mathcal{S}}^{cls}),
\end{equation}
where the first term is the mean square error between the feature maps $f_{\mathcal{T}}$ and $f_{\mathcal{S}}$, produced by the backbone module of the teacher and student models, respectively. The term $\mathcal{L}_{BCE}$ denotes the binary cross entropy loss. $y_{\mathcal{T}}^{cls}$ refers to the top-$k$ classification prediction after the temporal aggregation of the teacher model, and $y_{\mathcal{S}}^{cls}$ refers to that of the student model. $k$ is the number of proposals in the feature selection module before the temporal aggregation. We set $k=30$ following the default setting of YOLOV. We do not particularly compute the loss of objectiveness and bounding box prediction because they are unchanged in the temporal aggregation module.

\subsubsection{Spatial Refinement Stage (SRS).} 
TAM consists of two key components: a Feature Selection Module, which selects high-quality prediction proposals, and a Feature Aggregation Module, which fuses these proposals across multiple frames. However, due to the inconsistency between the training pipelines of the single-frame detection head (backbone) and the TAM, the TRS, which mostly follows the training setting of the TAM, may lead to suboptimal adaptation on the backbone side. Recognizing that the TAM can reliably improve prediction quality, we propose using the output class score of YOLOV, instead of YOLOX, in the teacher model as higher-quality pseudo labels to guide the fine-tuning of the detection head of YOLOX in the student model. In the SRS, only the backbone of the teacher model is updated via EMA, ensuring that the adaptation focuses on the spatial feature extraction process while leveraging the enhanced temporal information from the TAM. The loss is given as follows:
\begin{equation}
    \mathcal{L} = \mathcal{L}_{MSE}(f_{\mathcal{T}}, f_{\mathcal{S}}) + \mathcal{L}_{BCE}(y_{\mathcal{T}}^{cls}, y_{\mathcal{S}}^{cls}) + \gamma \mathcal{L}_{cls},
\end{equation}

\noindent where $\gamma$ is the weighting factor. The new loss term $\mathcal{L}_{cls}$ is the certainty-aware binary cross entropy loss between the filtered class score from the teacher and student model:
\begin{equation}
\begin{aligned}
    \mathcal{L}_{cls}  = -\frac{1}{N}\sum_{i}^{N} p^{i}_{\mathcal{S}} \Big[ \frac{1}{n_{c}} & \sum_{c}^{n_c}  \left( s_{\mathcal{T}}^{i,c} \log(s_{\mathcal{S}}^{i,c}) \right. \\
    & \left.  +(1-s_{\mathcal{T}}^{i,c}) \log(1-s_{\mathcal{S}}^{i,c}) \right) \Big],
\end{aligned}
\end{equation}
\noindent where $c$ is the index of the category, $n_c=30$ is the number of classes, and $i$ and $N$ are the index and number of detected objects in the sequence. $s_{\mathcal{S}}^{i,c}$ and $s_{\mathcal{T}}^{i,c}$ are the $i$-th output scores of class $c$ for the student and teacher models, respectively. $p^{i}_{\mathcal{S}} \in (0,1)$ is the normalized objectiveness score in the student model output, serving as the weight of the pseudo-label. It can be viewed as the certainty measurement of the object's existence; the greater $p^{i}_{\mathcal{S}}$ indicates the higher confidence of the particular pseudo label.
 
\subsubsection{Alternate Refinement.} 
STAR-MT training is periodical, with the TRS and SRS having identical iterations $\tau$ in each period. Given $k$ the index of the period, TRS is executed in iterations $[2k\tau, 2k\tau+\tau)$ and SRS in iterations $[2k\tau+\tau, 2k\tau+2\tau)$. During the experiment, it was observed that the order of those two stages only had a trivial impact on the overall performance.

Although early stopping is not explicitly implemented in our approach, we utilize the mean self-entropy \cite{li2021free} of the class score from the teacher model as a performance criterion for all output checkpoints. This mean self-entropy, denoted as $H$, serves as a measure of reliability for pseudo labels; a lower $H$ indicates greater confidence in the teacher model in guiding the student. The checkpoint corresponding to the minimal value of $H$ is selected as our output model. The formula to compute $H$ is as follows:
\begin{equation}
    H  = -\frac{1}{Nn_{c}} \sum_{i}^{N}  \sum_{c}^{n_c} s_{\mathcal{T}}^{i,c} \log(s_{\mathcal{T}}^{i,c}).
\end{equation}
\section{Experiment}

\begin{figure*}
    \centering
    \includegraphics[width = 0.99\linewidth]{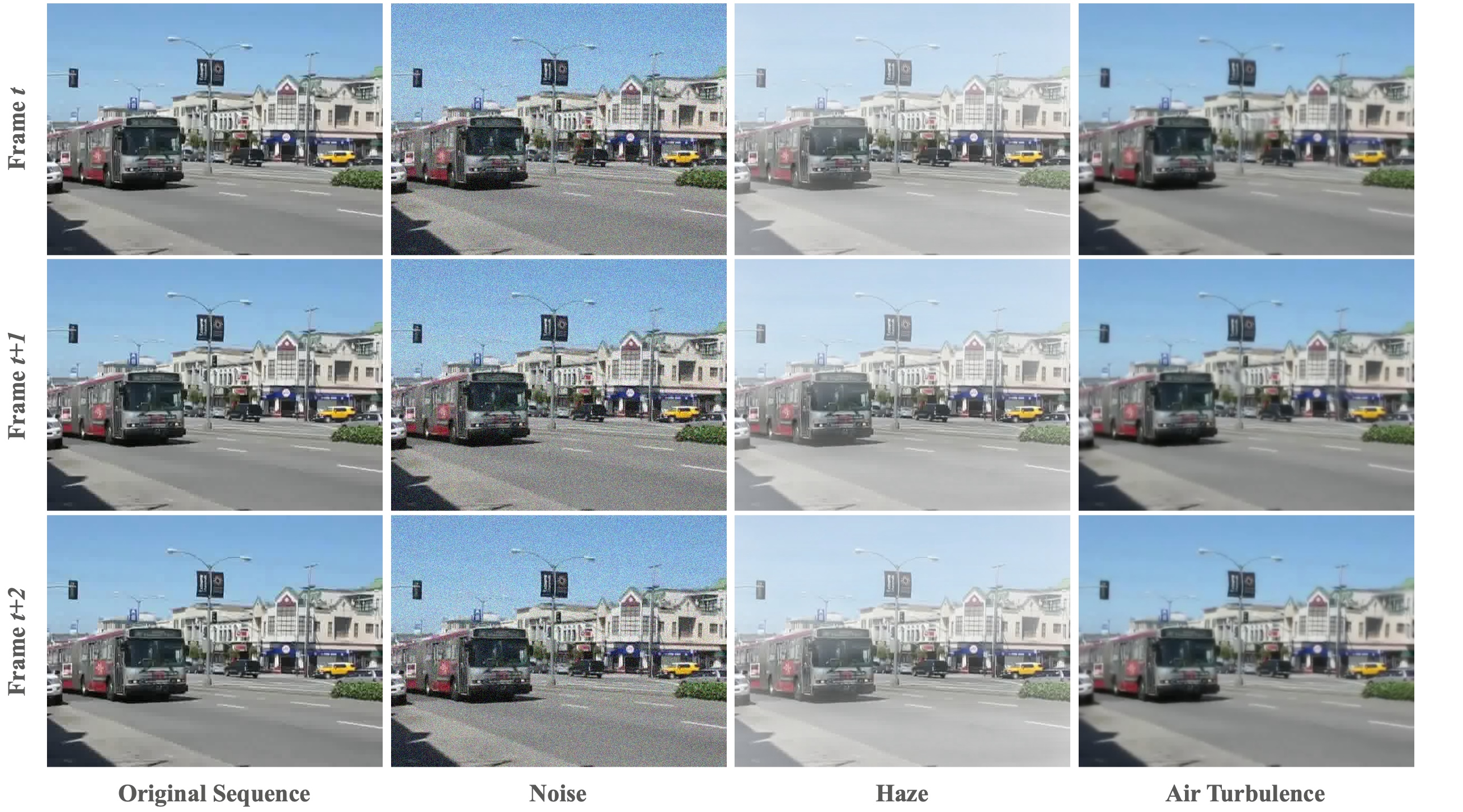}
    \caption{A snippet of the ImageNetVOD dataset and three forms of degradation. The original frames are from the testing video \textit{ILSVRC2015\_test\_00028000.mp4} and $t=32$.}
    \label{fig:test_sample}
\end{figure*}

\subsection{Adverse image condition synthesis}
\label{synthesis}
Real-world VOD often faces the challenge of domain gaps caused by adverse image conditions. Because of the complexity of image degradation, testing domain adaptation algorithms under various conditions is desired. However, appropriate datasets for testing the domain adaptation algorithm for VOD models trained on the ImageNetVOD dataset are unavailable. In this work, we synthesized videos in three common imaging conditions: noise, air turbulence, and haze. Each video has its distinct degradation parameter, with the profile varies temporally. A sample image sequence from the original dataset and the associated three degraded sequences are shown in Fig. \ref{fig:test_sample}. The unsupervised property of the algorithm guarantees that our method will be effective in real-world unknown degradations. The simulation of the three adverse image conditions is described below:

\noindent \textbf{Noise.} Noise is the predominant degradation in the low-light conditions. The noise in our experiment is modeled with:
\begin{equation}
    \widetilde{I}(h,w,t) = I(h,w,t) + n(h,w,t),
\end{equation}
\noindent where $I$ is the input image sequence, $\widetilde{I}$ is the degraded image sequence, $h$, $w$, and $t$ are the sequence's height, width, and frame indices. $n(h,w,t)\sim \mathcal{N}(0, \sigma^2)$ is the Gaussian noise. We randomly sample the variance of the noise in $\sigma^2\in [10/255, 50/255]$. Each sequence has its distinct variance.

\noindent \textbf{Air Turbulence.} In long-range imaging conditions, air turbulence may affect the performance of computer vision models significantly \cite{Liu_2024_WACV}. The air turbulence primarily causes random pixel displacement and spatially varying blur on the image \cite{chan2023computational}. We utilized the popular P2S simulator \cite{chimitt2020simulating, Mao_2021_ICCV} to synthesize the degraded video:
\begin{equation}
    \widetilde{I}(h,w,t) = \text{P2S}(I(h,w,t); n(h,w,t)).
\end{equation}
The P2S simulator converts the Gaussian random seed $n$ to spatially varying pixel displacement and blur. Inspired by \cite{chimitt2022real, zhang2022imaging}, we applied the temporal correlation and varying kernel size to improve the diversity and fidelity of the synthetic turbulence. Each image sequence has a distinct turbulence strength and profile.

\noindent \textbf{Haze.} Haze is a crucial adverse image condition in VOD application scenarios, especially in surveillance systems and automated driving. The hazy video can be modeled with the transmission function \cite{cai2016dehazenet}:
\begin{equation}
    \widetilde{I}(h,w,t) = I(h,w,t)e^{-\beta d(h,w,t)} + A(1-e^{-\beta d(h,w,t)} ),
\end{equation}
\noindent where $e^{-\beta d(h,w,t)}$ is the transmission rate, $\beta$ is the scattering coefficient, $A=255$ is the maximum intensity of a pixel, and $d(h,w,t)$ is the relative depth value measured by \cite{Godard_2019_mono}. Like other degradations, each sequence has its own scattering coefficient randomly sampled from a uniform distribution $\beta \in [0.5,1.5]$. 

\begin{table*}
\centering
\setlength{\aboverulesep}{0pt}
\setlength{\belowrulesep}{0pt}
\resizebox{0.99\textwidth}{!}{
\begin{tabular}{l|ccc|ccc|ccc}
\toprule[1pt]
    Degradation & \multicolumn{3}{c|}{Noise} & \multicolumn{3}{c|}{Air Turbulence} & \multicolumn{3}{c}{Haze}   \\
    \hline
Model & YOLOV-S & YOLOV-L & YOLOV-X & YOLOV-S & YOLOV-L & YOLOV-X & YOLOV-S & YOLOV-L & YOLOV-X \\
\hline
Source-only  & 38.0 & 57.0 & 60.4 & 63.2 & 72.7 & 73.9 & 57.2 & 70.7 & 73.0 \\
\hline
PL \textit{w.} SE \cite{li2021free}  & 47.8  & 61.4  & 62.3  & 64.3  & 73.2  & 74.1  & 61.1  & 73.2  & 74.5  \\
\hline
Basic MT & 54.8  & 70.3  & 70.6  & 64.1  & 74.2  & 74.4  & 64.7  & 75.3  & 76.9  \\
STAR-MT   & \textbf{57.6}  & \textbf{71.4}  & \textbf{71.5}  & \textbf{65.2}  & \textbf{75.0}  & \textbf{75.7}  & \textbf{68.1}  & \textbf{78.0}  & \textbf{78.9}  \\
\hline
Oracle    & 61.0  & 72.5  & 72.7  & 66.7  & 76.4  & 78.3  & 69.7  & 79.6  & 80.2  \\
\bottomrule[1pt]
\end{tabular}
}
\caption{Performance comparison on AP50(\%). The larger, the better. ``PL" refers to the pseudo-label method, and ``Source-only” refers to the models trained by only using labeled source domain data.}
\label{table:overall}
\end{table*}

\subsection{Dataset and baselines}
\begin{figure*}[t]
\small
\includegraphics[width=\linewidth]{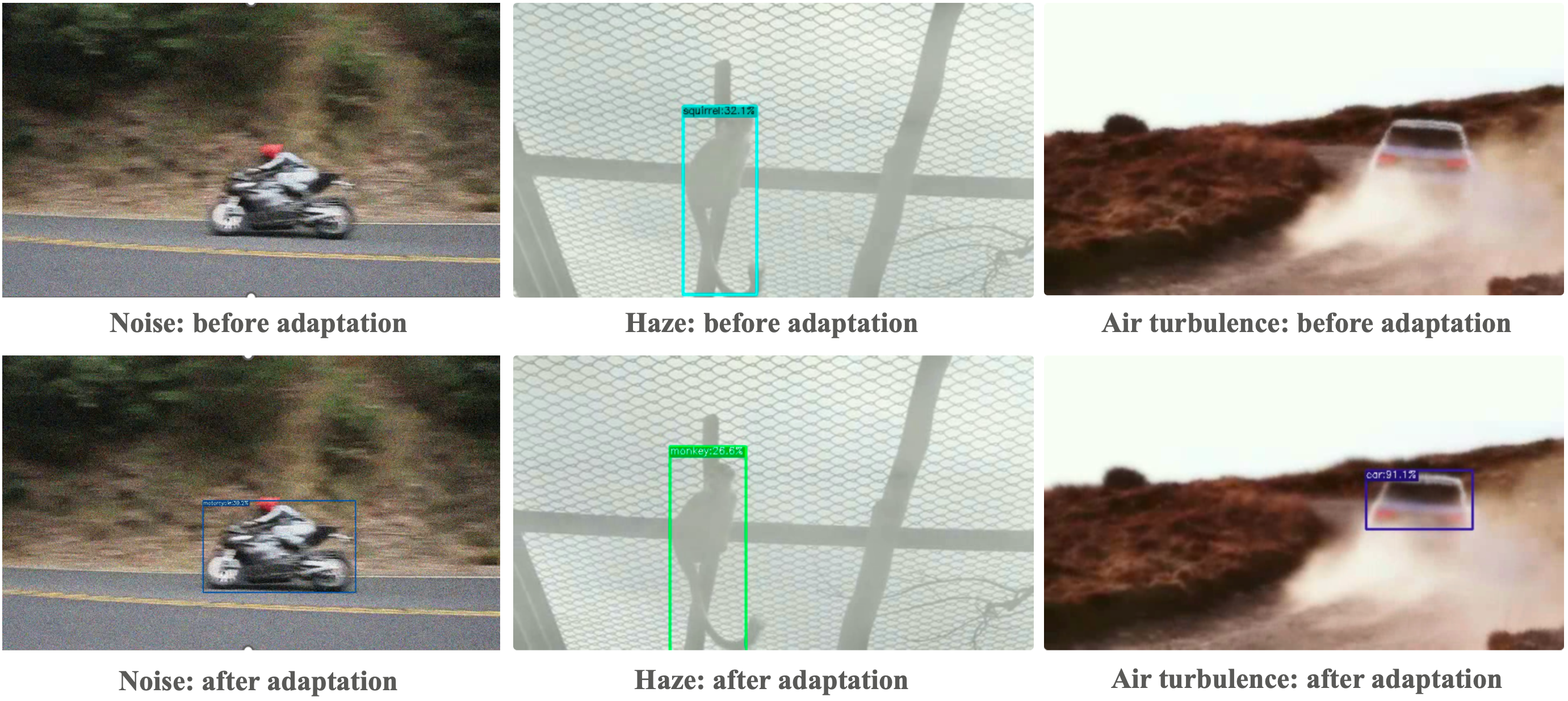}%
\caption{Visual comparison before and after the SFDA by STAR-MT. All experiments are conducted with YOLOV-S.}
\label{fig:effect}
\end{figure*}

In Section \ref{synthesis}, for each degradation type identified, we generated a corresponding synthetic target-domain dataset utilizing the ImageNetVID dataset \cite{russakovsky2015imagenet} as the source. Comprising 30 classes set against diverse natural backdrops, ImageNetVID provides over 1 million training frames and more than 100,000 validation frames. We used all frames from this source dataset to synthesize the target-domain datasets. Consequently, each domain can retain the same set of labels. Fig \ref{fig:effect} shows a snippet of the testing set of our synthetic degradations, along with the visualization of the detection results before and after the domain adaptation. The visual comparison proves the efficacy of the proposed method.

Following its publicly available codebase, we trained the YOLOV in all three scales — small (S), large (L), and extra-large (X) — using the source dataset. In our experiment, the post-processing method was omitted as it does not pertain to our algorithms. These models were then directly tested on target domains, with the findings detailed in Table \ref{table:overall}. The Average Precision at $50\%$ threshold (AP50) on the source domain is registered as $77.3\%$, $83.6\%$, and $85.0\%$ for YOLOV-S, YOLOV-L, and YOLOV-X, respectively. The significant degradation in performance, when we test the source-train model on the target domain dataset, indicates that challenging image conditions markedly reduce the performance of the VOD models.

In addition to the initial training, we used a supervised approach to fine-tune the source-trained YOLOV models on the target domain datasets. This supervised fine-tuning serves as a theoretical benchmark for the upper limit of performance achievable through unsupervised adaptation. Deviating from the original pipeline, which involves training the base detector prior to the temporal aggregation module, we discovered that directly fine-tuning the temporal aggregation module leads to improved outcomes. Therefore, our fine-tuning focuses solely on the temporal aggregation module for the target domains, as detailed in \cite{shi2023yolov}. The results of this supervised fine-tuning, labeled as ``oracle", are also presented in Table \ref{table:overall}.

Before implementing the mean-teacher-based methods, we conducted a preliminary experiment with the pseudo-label (PL) algorithm. In this approach, models trained on the source domain are employed to process all videos in the target domain's training set, generating initial predictions. They are then filtered by threshold 0.5 on the product of objectiveness and the maximal class scores to generate pseudo labels. Since fine-tuning the single-frame backbone always causes catastrophic failure, we fixed the parameters in the backbone module and only trained the temporal aggregation module with pseudo labels. After training, we utilized the self-entropy \cite{li2021free} as the indicator to select the potential best model. The result is also demonstrated in Table \ref{table:overall}.

\subsection{Implementation details of STAR-MT}
Adhering closely to the YOLOV codebase, we maintained most of the original settings unaltered. For hyperparameter configuration, we empirically set the teacher model's smoothing coefficient, $\alpha$, to 0.9995 and the weighting factor, $\gamma$, of the $\mathcal{L}_{cls}$ to 0.2. The model training was executed using Stochastic Gradient Descent (SGD) with a batch size of 1 over 10,000 iterations. We initialized the learning rate at $2 \times 10^{-4}$ and applied a cosine annealing scheduler, tapering it down to $1 \times 10^{-4}$. In the evaluation phase, only the teacher model was utilized for inference. The mean Average Precision (mAP) was calculated with an IoU threshold of 0.5. All experiments are conducted on NVIDIA 3080 Ti and V100 GPUs.

\begin{figure}[t]
\small
\begin{minipage}[b]{.49\linewidth}
  \centering
\includegraphics[width=\linewidth]{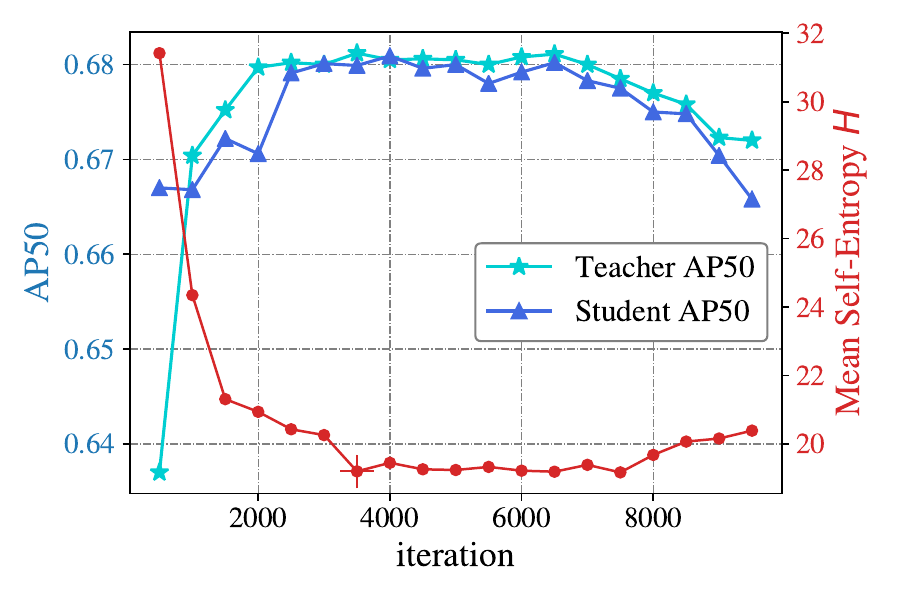}%
  \vspace{0.1cm}
  \centerline{(a) YOLOV-S}
\end{minipage}
\hfill
\begin{minipage}[b]{0.49\linewidth}
  \centering
\includegraphics[width=\linewidth]{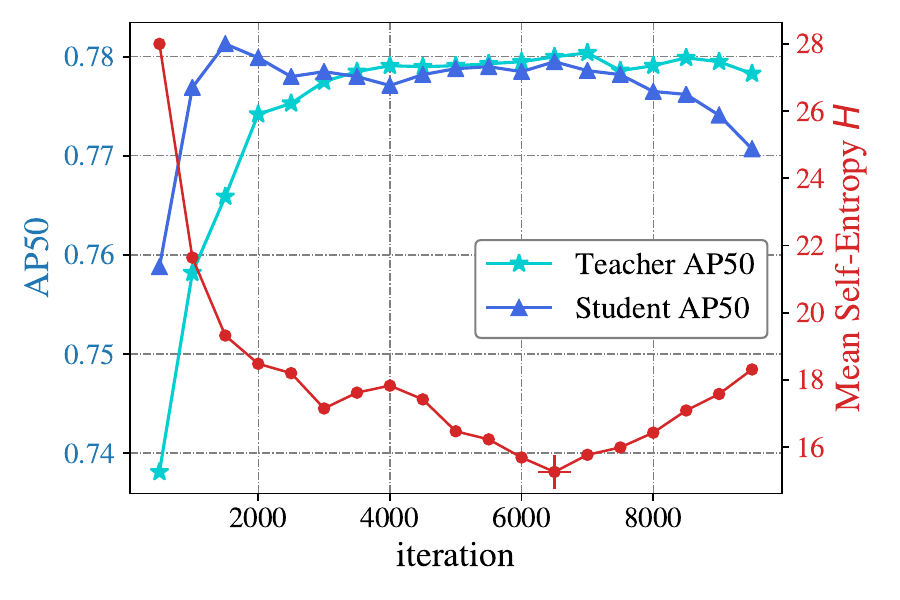}%
  \vspace{0.1cm}
  \centerline{(b) YOLOV-L}
\end{minipage}
\caption{The teacher model's AP50 and mean self-entropy $H$ variation in the STAR-MT training of YOLOV-S and YOLOV-L. Both experiments are conducted on clean $\rightarrow$ haze. The $H$ indicating the best teacher model are marked in the figures with ``\textcolor{red}{+}".}
\label{fig:curve}
\end{figure}

For each sequence in our domain adaptation experiments, 32 frames are loaded. Mosaic augmentation was disabled for all these experiments. However, we have retained both random flip and perspective transformations, applying these consistently to both weakly and strongly augmented sequence pairs. The key distinction between weak and strong augmentation lies in the strength of random chromatic transformation. Random erasing is involved only in the strong augmentation. In the TRS, random masking is applied to restrict the temporal information the student model can access, compelling it to enhance the temporal aggregation capability. The masking rate is $r\%$ where $r$ is randomly sampled from $[0, 75]$.

The performance of the STAR-MT method is demonstrated in Table \ref{table:overall}. Our method shows a significant improvement in the SFDA for VOD under all three adverse image conditions. It also demonstrates a clear advantage over conventional methods like pseudo-labeling and basic mean-teacher learning. Notably, although the method seems straightforward and not complicated, \textit{the performance of our method closely approaches that of supervised fine-tuning}. To illustrate the correlation between mean self-entropy $H$ and the performance of SFDA, we drew the variations in model performance alongside the change in $H$ value on the evaluation set, as shown in Fig \ref{fig:curve}. The $H$ values were calculated using a sliding window average over 100 iterations. We can observe the lowest value of $H$ aligns well with the peak performance of the model.

\subsection{Ablation study}
\noindent \textbf{Efficacy of alternate refinement.} One major novelty in this paper is the spatial refinement stage, as an alternately updated module in addition to the normal mean teacher learning framework. The key insights behind this are 1) temporally enhanced features of the teacher model can be used to generate reliable pseudo labels for the training of the single-frame detection head, and 2) training the single-frame detection head under the YOLOV setting is suboptimal. Thus, it needs additional guidance. From the comparison between the basic mean-teacher method (TRS only) and the proposed STAR-MT in Table \ref{table:overall}, we can verify the efficacy of alternate refinement with the spatial refinement stage. To demonstrate that the pseudo labels from the YOLOV are of higher quality, we conducted source-free domain adaptation for the YOLOX. We follow the mean-teacher framework and use the pseudo labels generated by YOLOX and YOLOV to guide the training of the student network. The result is shown in Table \ref{tab:SRS}. The model guided by the temporally refined labels in YOLOV gets better performance, which provides evidence for the efficacy of SRS.

\begin{table}
    \centering
    \begin{tabular}{l|c|c}
    \hline
        Model & YOLOX-S &  YOLOX-L  \\
        \hline
        Source-only  & 35.9 & 56.6  \\
        PL guided by YOLOX  & 49.2 & 61.3  \\
        PL guided by YOLOV & 51.0 & 62.9\\
        \hline 
        Oracle & 56.7 & 66.0 \\
    \hline
    \end{tabular}
    \caption{The efficacy of YOLOV as the teacher model for the SFDA of the single-frame detection backbone. All experiments are conducted on clean $\rightarrow$ noise. The metric is AP50(\%), the larger, the better.}
    \label{tab:SRS}
\end{table}

\begin{table}
    \centering
    \begin{tabular}{c|c|c|c|c}
    \hline
        $\mathcal{L}_{MSE}$ & $\mathcal{L}_{BCE}$ & $\mathcal{L}_{cls}$ & YOLOV-S  & YOLOV-L \\
        \hline
        \xmark & \cmark & \cmark & 62.1 & 71.5 \\
        \cmark & \cmark & \xmark & 67.6 & 77.4 \\
        \cmark & \xmark & \cmark & 67.8 & 77.3 \\
        \cmark & \cmark & \cmark & \textbf{68.1} & \textbf{78.0} \\
    \hline
    \end{tabular}
    \caption{The efficacy of losses. All experiments are conducted on clean $\rightarrow$ haze. The metric is AP50(\%), the larger, the better.}
    \label{tab:losses}
\end{table}

\noindent \textbf{Efficacy of losses.} In our study, the three utilized loss functions are categorized as feature alignment loss $\mathcal{L}_{MSE}$ and pseudo-label based losses ($\mathcal{L}_{BCE}$ and $\mathcal{L}_{cls}$). We experimented with various reasonable combinations of these loss terms to assess their impact. In all combinations, at least one pseudo-label-based loss was maintained. The results are detailed in Table \ref{tab:losses}. Initially, we excluded the feature alignment loss $\mathcal{L}_{MSE}$ and observed a significant decline in adaptation performance. This indicates the model's high sensitivity to label quality and the importance of restricting the feature space generated by the detection head. Further, we excluded $\mathcal{L}_{MSE}$ and $\mathcal{L}_{cls}$ separately to evaluate their individual contributions. The results confirmed the effectiveness of both losses.

\begin{table}
    \centering
    \begin{tabular}{c|c|c|c|c}
    \hline
        Model & $\tau = 50$ & $\tau = 100$ & $\tau = 200$  & $\tau = 500$ \\
        \hline
        YOLOV-S & 68.0 & \textbf{68.1} & 67.6 & 67.8 \\
        YOLOV-L & 76.9 & 77.5 & \textbf{78.0} & 77.7 \\
        YOLOV-X & 78.2 & 78.4 & \textbf{78.9} & 78.8 \\
    \hline
    \end{tabular}
    \caption{The impact of the number of iterations in each stage. All experiments are conducted on clean $\rightarrow$ haze. The metric is AP50(\%), the larger, the better.}
    \label{tab:iteration}
\end{table}

\noindent \textbf{Number of iterations for each stage.} We also evaluated the optimal number of fine-tuning iterations, $\tau$, for each stage within a period. For this purpose, we conducted experiments that set $\tau$ to various values: 50, 100, 200, and 500, while maintaining 10,000 training iterations. The determination of optimal results within this range was based on the mean self-entropy $H$ values, where the teacher model associated with the first local minima of $H$ is selected. This assessment was carried out across all three model scales, and the findings are presented in Table \ref{tab:iteration}. The results indicate that different scales of the model may require distinct hyperparameters to achieve optimal adaptation performance.
\section{Conclusion}

In this paper, we propose a pioneering approach to explore the source-free domain adaptation (SFDA) for video object detection (VOD). Specifically, we developed a novel SFDA method for a one-stage-based detector, YOLOV. The proposed STAR-MT technique significantly improves the performance of the video object detector in adverse image conditions without access to the target domain label or source domain data. Owing to its unsupervised nature, this work can be seamlessly applied to real-world scenarios requiring VOD models. The proposed method could serve as a baseline for future research in unsupervised domain adaptation for video object detection.

\section{Acknowledgment}
Futurewei Technologies Inc. funded this research while the
first author was an intern there. We are grateful to the company’s
IC Lab for the research assistance.

{
    \small
    \bibliographystyle{ieeenat_fullname}
    \bibliography{main}
}


\end{document}